%% file: main.tex
\documentclass[letterpaper]{article}
\usepackage{spconf,amsmath,amssymb,graphicx}
\usepackage{xcolor}
\usepackage{algorithm}
\usepackage{algpseudocode}
\usepackage{mdframed}
\usepackage{subcaption}  

\usepackage[textsize=scriptsize]{todonotes}

\usepackage{hyperref}
\usepackage{booktabs}
\usepackage{graphicx}
\hypersetup{
    colorlinks=true,
    linkcolor=black,
    urlcolor=black,
    citecolor=blue,
}
\usepackage{microtype}

\usepackage{inconsolata}
\usepackage{amssymb,amsmath,amsthm}
\theoremstyle{definition}

\numberwithin{equation}{section}
\newcommand{\va}{\mathbf{a}}
\newcommand{\vb}{\mathbf{b}}
\newcommand{\vc}{\mathbf{c}}
\newcommand{\vx}{\mathbf{x}}
\newcommand{\vy}{\mathbf{y}}

\newcommand{\mX}{\mathbf{X}}
\newcommand{\mY}{\mathbf{Y}}
\newcommand{\mW}{\mathbf{W}}
\newcommand{\mT}{\mathbf{T}}

\newcommand{\mypar}[1]{{\bf #1.}}

\title{Single-core Superscalar Optimization of \\Clifford Neural Layers}
\name{Xuanqiang `Angelo' Huang$^{*}$\thanks{Authors declare equal contribution}, Ruben Ciranni$^{*}$, Giovanni Spadaccini$^{*}$, Carla J. L\'opez Zurita$^{\dagger}$   
}

\address{
\vspace{0.5em}
\texttt{\{\href{mailto:hxuanqiang@ethz.ch}{hxuanqiang}, \href{mailto:rciranni@ethz.ch}{rciranni}, \href{mailto:gspadaccini@ethz.ch}{gspadaccini}, \href{mailto:calopez@ethz.ch}{calopez}\}@ethz.ch} \\ 
ETH Z{\"u}rich\\
		 $^*$Department of Computer Science, $^{\dagger}$Department of Mathematics
} 
\begin{document}
%
\maketitle
\hypersetup{urlcolor=blue}


\begin{abstract}
Within the growing interest in the physical sciences in developing networks with equivariance properties, Clifford neural layers shine as one approach that delivers $E(n)$ and $O(n)$ equivariances given specific group actions.
In this paper, we analyze the inner structure of the computation within Clifford \textit{convolutional} layers and propose and implement several optimizations to speed up the inference process while maintaining correctness. In particular, we begin by analyzing the theoretical foundations of Clifford algebras to eliminate redundant matrix allocations and computations, then systematically apply established optimization techniques to enhance performance further.
We report a final average speedup of 21.35x over the baseline implementation of eleven functions and runtimes comparable to and faster than the original PyTorch implementation in six cases. In the remaining cases, we achieve performance in the same order of magnitude as the original library.

\vspace{.5em}
\includegraphics[width=1.25em,height=1.15em]{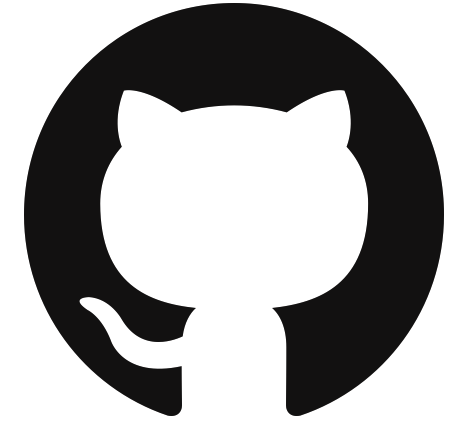}\hspace{.75em}
\parbox{\dimexpr\linewidth-7\fboxsep-7\fboxrule}{\url{https://github.com/Flecart/superscalar-clifford-neural-layers}}
\vspace{-1em}
\end{abstract}

\section{Introduction}\label{sec:intro}

\mypar{Motivation} Clifford neural networks are a promising family of models that leverage Clifford algebras~\cite{dorstGeometricAlgebraComputer2009,ruheCliffordGroupEquivariant2023,ruheGeometricCliffordAlgebra2023} for predictions with built-in invariance for common transformations such as translation, rotations and scaling, i.e. $O(n)$ and $E(n)$ transformations.
These symmetries are particularly common in the physical sciences, where many real-world phenomena are intrinsically governed by geometric relationships and symmetries~\cite{atiyahGeometryPhysics2010,iachelloSymmetrySearchOrder2011}.
This innovation has spurred many applications; examples include partial differential equation (PDE) modeling~\cite{brandstetterCliffordNeuralLayers2023}, protein structure prediction~\cite{pepeCliffordGroupEquivariant2024}, and prediction of interatomic potentials~\cite{batznerE3equivariantGraphNeural2022}.
One advantage of Clifford networks above other attempts to imbue networks with equivariance properties~\cite{kondorClebschGordanNets2018,weiler3DSteerableCNNs2018,fuchsSE3Transformers3DRotoTranslation2020} is their \textit{implicit} representation of the equivariances given some specific group action, without the need of \textit{explicitly} designing these properties, which often involve specific mathematical tools such as the Clebsch-Gordan product~\cite{kondorClebschGordanNets2018,liUnifyingO3Equivariant2024}






Modern Geometric Clifford Algebra Networks (GCANs) are used to model three-dimensional rigid body transformations and to simulate large-scale fluid dynamics, demonstrating significant performance improvements over traditional methods~\cite{ruhe2023geometriccliffordalgebranetworks}.
Building on the success of integrating theoretical geometric priors into neural networks, these layers extend traditional convolutional layers by incorporating real Clifford algebras~\cite{dorst2009geometric}.

The work presented in this paper is an \textit{architecturally specific inference optimization} of work published by Ruhe et al.~\cite{ruhe2023geometriccliffordalgebranetworks}, where they propose a PyTorch~\cite{paszkePyTorchImperativeStyle2019} based implementation of GCANs, compatible for both CPU and GPU.  While our focus is on optimizing for single-core CPU inference using full-precision arithmetic and AVX2 SIMD instructions, the optimization strategies we propose are transferable to any other modern superscalar hardware architecture. We look, in particular, into the integration of the Clifford Kernel with the corresponding linear operation in the Clifford layers.




\mypar{Contributions}
We describe our contribution as follows: 
(1) We introduce fully compatible and optimized Python bindings for the most important Clifford functions, including 1D, 2D, and 3D Clifford linear forward functions, as well as their equivalents for convolutional layers, activation functions, and g3 convolutional and g3 transpose layers, totalling 11 functions.
(2) We provide a simple and organised infrastructure to test both the correctness and performance of various versions of the Clifford layers. We find and fix a bug in the original repository (see \href{https://github.com/microsoft/cliffordlayers/pull/17}{here}).
(3) We optimize the initial C implementation by leveraging ILP (Instruction Level Parallelism)~\cite{bryant2011computer}, theory-based optimizations grounded in the fundamental properties of Clifford algebra's additions and multiplications, and vectorize all of the functions using Intel AVX2 intrinsics~\cite{DetailsIntelAdvanced}.

Our implementation achieves comparable performance to the original PyTorch version across many functions, occasionally surpassing it, while preserving the same level of accuracy and correctness during tests.


\section{Background}\label{sec:background}

\mypar{Geometrical deep learning} Geometrical deep learning draws inspiration from Felix Klein's Erlangen Program in geometry to provide \textit{common mathematical tools} to analyze the most popular architectures~\cite{bronsteinGeometricDeepLearning2021a}. In recent years, numerous neural networks have been developed based on principles of symmetry, invariance, geometric priors, and group actions, such as Graph Neural Networks~\cite{wuComprehensiveSurveyGraph2021a}, Spherical and Steerable Convolutional Neural Networks~\cite{cohenSphericalCNNs2018}, and Equivariant Neural Networks~\cite{gerkenGeometricDeepLearning2021}.
Clifford neural networks represent a specific approach to embed symmetries, such as the orthogonal transformations $O(n)$ and Euclidean motions $E(n)$, by leveraging algebraic structures.

\mypar{Real Clifford algebras}
Recall that an algebra over a field consists of a vector space $V$ over a field $\mathbb{F}$ equipped with a bilinear product $V \times V \rightarrow V$. \\
A real $n$-dimensional Clifford algebra $Cl_{p,q}(\mathbb{R})$ is an associative algebra generated by an orthonormal basis $\{e_1, \dots, e_n\}$ of $n = p + q$ vectors in $\mathbb{R}^n$ subject to the following relations which define how the bilinear product of the algebra operates on the generating orthonormal basis:
\begin{align}
e_i^2 = -1 \quad &\text{for } 1 \le i \le p \\
e_i^2 = +1 \quad &\text{for } p < i \le p + q = n \\
e_i e_j = - e_j e_i \quad &\text{for } i \ne j
\end{align}
These relations define the basis of the vector space $G^n$ of the Clifford algebra (note that this is different from the generating vector space), which is given by:
\[\{1, e_1, e_2, \dots, e_{n},\ e_1 e_2,\ \dots,\ e_{n-1} e_{n},\ \dots,\ e_1 e_2 \cdots e_{n}\}.\] 
Note that this basis of $G^n$ has dimension $2^n$, and we say that $2^n$ is the number of blades of the algebra. Moreover, we define the grade of each element of the basis of $G^n$ as the number of elements of the generating basis that appear in it. For example, the basis elements of $\{1, e_1, e_2, e_1e_2\}$ have grade $\{0, 1, 1, 2\}$. Linear combinations of basis elements with the same grade $k$ are known as $k$-vectors, while elements of $G^n$ are multivectors. For example, 3 is a scalar, $e_1 + 2e_2$ is a vector, $4e_1e_2$ is a bivector, while $e_1 + 2e_1e_2$ is a multivector. \\
The bilinear product of the Clifford algebra satisfies the following relations:
\begin{align}
\va \vb \in G_{p+q} \\
(\va\vb)\vc = \va(\vb\vc) \\
\lambda\va = \va\lambda \\
\va(\vb + \vc) = \va\vb + \va\vc
\end{align}
where $\mathbf{a}, \mathbf{b}$ belong to the vector space $G^{p+q}$ of $Cl_{p,q}(\mathbb{R})$ and $\lambda$ is a scalar.

Being $G^{p+q}$ a vector space, the sum between multivectors is computed element-wise, while we can obtain closed-form formulas for the bilinear product in $Cl_{p,q}(\mathbb{R})$ by collecting terms that multiply the same basis elements. For example, for $\va = a_0 + a_1e_1 + a_2e_2 + a_{12}e_1e_2$ and $\vb = b_0 + b_1e_1 + b_2e_2 + b_{12}e_1e_2$ multivectors of $Cl_{2,0}(\mathbb{R})$, we have:
\begin{align}
\va + \vb &= \begin{aligned}[t]
    &(a_0 + b_0) \cdot 1 \\
    &+ (a_1 + b_1)\, e_1 \\
    &+ (a_2 + b_2)\, e_2 \\
    &+ (a_{12} + b_{12})\, e_1 e_2
\end{aligned} \\
\va \vb &= \begin{aligned}[t]
    &(a_0 b_0 + a_1 b_1 + a_2 b_2 - a_{12} b_{12}) \cdot 1 \\
    &+ (a_0 b_1 + a_1 b_0 - a_2 b_{12} + a_{12} b_2)\, e_1 \\
    &+ (a_0 b_2 + a_1 b_{12} + a_2 b_0 - a_{12} b_1)\, e_2 \\
    &+ (a_0 b_{12} + a_1 b_2 - a_2 b_1 + a_{12} b_0)\, e_1 e_2
\end{aligned}
\end{align}

It is worth noticing that $Cl_{1,0}(\mathbb{R})$ is isomorphic to complex numbers $\mathbb{C}$ while $Cl_{2,0}(\mathbb{R})$ is isomorphic to quaternions $\mathbb{H}$. We use this fact to test the correctness of our implementation; for example, we compare the result of a $Cl_{1,0}(\mathbb{R})$ layer with that of a complex layer, which is already implemented in most deep learning libraries.

\mypar{Clifford neural layers} Clifford neural layers are a natural extension of deep learning layers like linear layers, convolutions, and activations, which act on Clifford algebra tensors instead of real tensors. For example, a Clifford linear layer is defined as follows:
\begin{equation}
    \vy_i = \vb_i + \sum_{j = 1}^{n} \va_{ij} \vx_j \quad \text{for all }  i \in \{1, \dots, M\}
\end{equation}
Where $\mW \in {G^{p + q}}^{M \times N}$, $\vx \in {G^{p + q}}^{N}$, $\vb, \vy \in G_{p + q}^{M}$ and sum and multiplication symbols here represent Clifford sum and bilinear product as defined earlier.



\mypar{Single-core optimizations on superscalar architectures}\label{sec:background-lapack}
Most modern processors allow the execution of different instructions in parallel through different execution ports~\cite{johnsonSuperScalarProcessorDesign1989}, yet not all software is written to exploit these hardware features. Historically, LAPACK/BLAS~\cite{angersonLAPACKPortableLinear1990} and its theoretical counterpart ATLAS~\cite{whaleyAutomaticallyTunedLinear1998} have stood out as automatic, fast, and portable frameworks for sculpting software that is able to harness the underlying optimizations such as instruction level parallelism, out-of-order execution or register renaming.
However, the above frameworks are highly optimized for general matrix and vector operations, not for more specific computations. This opened the space to the development of other software, such as FFTW~\cite{frigoDesignImplementationFFTW32005,frigoFFTWAdaptiveSoftware1998} for Fourier transforms to attack different yet important computations by leveraging its mathematical structure. We argue in this paper that by applying other domain-specific optimizations for Clifford convolutional layers, we can reach a performance comparable or superior to the automatic frameworks cited above.

\begin{figure}[!ht]
    \centering
    \includegraphics[width=0.9\linewidth]{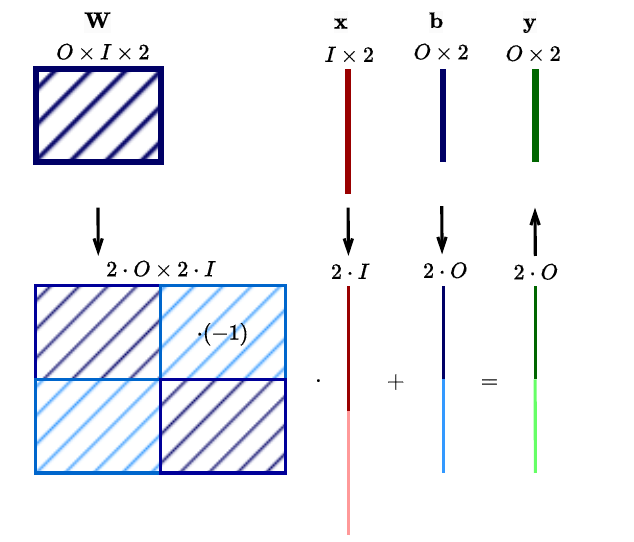}
    \caption{A visualization of matrix-vector multiplication in $Cl_{1, 0}(\mathbb{R)}$ using Clifford kernel technique employed in the original computation with many optimization pathways. See section~\ref{par:original-computation} for a discussion.}
    \label{fig:algo_clifford_linear}
\end{figure}

\section{Methods}\label{sec:yourmethod}
In this section, we describe the existing implementation of Clifford neural layers and the optimizations we perform.

\subsection{Original Implementation}
A PyTorch implementation of the Clifford neural layers is provided by the authors in \cite{brandstetter2023cliffordneurallayerspde}. In this implementation, a Clifford tensor $\mT \in G_{p+q}^{D_1, \times, \dots, \times D_d}$ is stored as a real tensor $G_{p+q}^{D_1, \times, \dots, \times D_d \times 2^{p+q}}$, where the last dimension corresponds to the components of each multivector entry. Using carefully thought-out transformations of input, parameters, and output, the built-in PyTorch layers for real numbers are used to produce the desired behaviour of the Clifford layer. This involves (1) permuting and flattening the dimensions of input and parameters, (2) transforming the weights into a custom Clifford kernel, and (3) permuting back into the original shape the output (see Figure~\ref{fig:algo_clifford_linear} for a detailed explanation of how this is achieved for matrix-vector multiplication in $Cl_{1, 0}(\mathbb{R})$). We provide a detailed algorithm for the Clifford linear layer for $Cl_{1, 0}(\mathbb{R})$ algebra for the sake of simplicity (see Algorithm~\ref{app:algo_clifford_linear}), but the same technique can be extended to all Clifford algebras and other types of layers.
\input{algorithms/cliffordlinearbasline}

\mypar{The Clifford kernel trick} \label{par:original-computation}
Algorithm~\ref{app:algo_clifford_linear} uses the Clifford kernel trick. This is a general technique used throughout the existing PyTorch implementation of Clifford neural layers. As an example, we examine how matrix-vector multiplication is achieved, while a similar technique is used for convolution. Figure~\ref{fig:algo_clifford_linear} depicts the computation: (1) $\mathbf{W}$ is transformed to the corresponding Clifford kernel (dark blue quadrants correspond to scalar entries of $\mathbf{W}$, light blue ones correspond vector entries), observe that upper right quadrant of the kernel is multiplied by \textcolor{purple}{-1}, this comes from the bilinear product formula of the $Cl_{1,0}(\mathbb{R})$ algebra: $\va \vb = (a_0b_0 \textcolor{purple}{-} a_1b_1) \cdot 1 + (a_0b_1 + a_1b_0)\,e_1$, (2) $\vx$ is reshaped and flattened as depicted (dark red line corresponds to scalar entries of the vector, light red one corresponds to vector entries), (3) standard matrix-vector multiplication is performed between Clifford kernel and reshaped $\vx$, this produces a vector in which the first half contains the scalar entries of the output and the second half contains the vector entries, (4) the reshaped and flattened bias is added to the result, (5) the result is reshaped back in the original form.
\subsection{Baseline C Implementation}\label{subsed:analysisbaseline}
We implement the baseline C functions strictly following the PyTorch code. We define four classes of functions that have similar characteristics: (1) Clifford kernels, (2) Clifford activation functions, (3) Clifford linear layers and (4) Clifford convolutional layers.

\mypar{Clifford kernels}
The authors of the Clifford Neural Layers~\cite{ruhe2023geometriccliffordalgebranetworks} implement the Clifford operations by \textit{expanding} the original layer weights into a new matrix named Clifford kernel that is \textit{specifically} tiled to reduce the problem of Clifford matrix multiplication or convolution to a standard matrix multiplication or convolution in $\mathbb{R}$.
This is depicted in Figure~\ref{fig:algo_clifford_linear} for $Cl_{1,0}(\mathbb{R)}$ matrix-vector multiplication; the same idea is used for convolutions.
With this technique, they were able to leverage the accelerated LAPACK libraries~\cite{angersonLAPACKPortableLinear1990} used underneath PyTorch. However, this approach incurs high memory and operational costs due to the need to replicate the weight matrix during kernel creation: for a Clifford algebra $Cl_{p,q}(\mathbb{R})$ with $p + q = n$, i.e. with $N = 2^n$ blades, each creation of a Clifford kernel incurs in $N$ times the space of the original weights and an additional number of operations in the order of $\mathcal{O}(\text{size})$ where $\text{size}$ is the total size of the weight matrix. Moreover, the reshapes needed for this technique make the underlying data not contiguous in memory, which adds additional overhead.

\mypar{Clifford activations} We notice that all operations on Clifford activations use $B \cdot N \cdot L_1 \cdot L_2$ additions to compute a cumulative quantity and $M$ multiplications to compute the final product operation for a total approximation of $\mathcal{O}(M)$ flops in the computation without considering division for the mean activation and other operations upper bounded by the same asymptotic analysis. Similarly, each computation needs to write to all elements of the array, making the lower bound of the memory transfers $\Omega(M)$. From this, we conclude the upper bound on the operational intensity of the computation is $1$, a result we verified empirically with roofline analysis in Section~\ref{sec:exp}.

\mypar{Clifford linear layers} All linear layers are standard matrix-vector operations where the summation and multiplication are defined on Clifford algebras; see Section~\ref{sec:background}. We use variable names defined in Table~\ref{tab:parameters_act_linear}. We report $4B \cdot O \cdot I$ multiplications, and $4B \cdot O \cdot I + 2B\cdot O$ additions on the \texttt{baseline} implementation in addition to data movements in the order of $\mathcal{O}(O \cdot I)$ for the computation of the kernel. Therefore, the total flops count is in the order of $\mathcal{O}(B \cdot O \cdot I)$. We need to access every weight matrix, input and output array at least once, leading to a lower bound of memory transfers of $B \cdot O \cdot N + B\cdot I \cdot N + N \cdot O\cdot I$. Observe that the working set of linear layers does not consider the batches, which means all matrices fit into the L1 cache for common values of $O$ and $I$, which usually do not extend over the thousands. 

\mypar{Clifford convolutional layers} Similarly, from the above analysis, by scaling only batch size, we can safely assume that a single batch element fits within the L1 cache, implying we will not have any further eviction after the initial cold miss. In this case, we have an upper bound of $\mathcal{O}(B \cdot CO \cdot CI \cdot L \cdot K)$ where $B, CO, CI, K$ are defined as in Table~\ref{tab:parameters_conv} and $L$ is the total size of the remaining dimensions.


\subsection{Optimizations}\label{sec:optimizations}
By using a custom C backend with the corresponding forward functions, we were able to leverage new and classical techniques that boost the performance of the executables. See section~\ref{sec:background-lapack} for an overview of classical techniques.

\input{algorithms/cliffordlinearoptimized}

\mypar{Clifford kernel inlining}
We replaced the Clifford kernel technique by computing the Clifford algebra operations directly in our specialized backend (linear, convolution, transposed convolution). This solves the disadvantages of the Clifford kernel technique mentioned in Section~\ref{subsed:analysisbaseline}. However, this prevents us from using existing linear algebra backends like LAPACK. We provide Algorithm~\ref{app:algo_optimized_clifford_linear}, which corresponds to the Clifford kernel inlining optimization of Algorithm~\ref{app:algo_clifford_linear}.
The original loop over the dimensionality of the Clifford algebra was inherently unrolled and, at the same time, induced some instruction-level parallelism (ILP) into the computations by using the distributivity of the bilinear product (the sum of products between all the combinations of blades are computed independently). The distributive property also allows us to write the computation in a way that consists almost solely of independent fused multiply adds. Moreover, we notice that increasing the number of blades of the algebra quadratically increases the number of independent fused multiply adds in the innermost loop (we have $|\{0, \dots, N - 1\}^2| = N^2$ independent fused multiply adds in the innermost loop in an algebra with $N$ blades), for this reason we later avoid loop unrolling for larger values of $N$ as, for example, already in a 2-dimensional Clifford algebra we have $(2^2)^2 = 16$ independent fused multiply adds in the innermost loop.

\mypar{Standard optimizations}
At the same time, we explored some loop reordering to achieve better read patterns and avoid cache misses.
We also explored some further \textit{unrolling}, looking to reduce computations and improve single-strided accesses.
We conducted autotuning by generating versions of the unrolled code parameterized by a variable. This allowed us to create multiple variants of the same function, each with a different unrolling factor. We then evaluated their performance and selected the best-performing version. In most cases, the optimal unrolling factor was 2 or 4, primarily for batch size. Larger unrolling factors generally led to increased memory access overhead, which negatively impacted performance.



\mypar{Vectorization}\label{sec:vecorization}
Moving on to vectorization, we used AVX2 intrinsics, allowing the computation of 8 single precision floating point operations at a time.
Functions with contiguous, strided accesses, such as linear layers, were straightforward to vectorize.
Convolution required more careful consideration because we allowed the implementation to support both kernel dilation and stride. As a result, some convolution functions resulted in non-contiguous data loads and stores, which was the case for convolution
layers in 1D, 2D and g3 2D normal and transposed convolution.
This generality limited the potential optimizations that could have been applied if these parameters were fixed.

\mypar{Other optimizations}\label{par:other-optimizations}
Specifically for VSiLU Clifford activation functions, we experimented with strength reduction by using approximations of the sigmoid function. This did not yield a significant improvement in speed-up, especially for large input sizes, so we decided to exclude this from our optimizations, also considering the approximation errors it induced.
We also experimented with different compilers and compiler flags, and we report the setup used for our benchmarks in Section~\ref{sec:exp} for more details.

\begin{table}
    \centering
    \begin{tabular}{@{}lcccc@{}}
        \toprule
        Layer  & Input (I) & Output (O) & Batch (B) & Blades (N)\\
        \midrule
        Linear      & 100       & 100        & $n$  &   2,4,8  \\
        \bottomrule
    \end{tabular}
    \caption{Parameters for linear layers, $n$ is input size parameter.
    Number of blades depends on dimensionality of the Clifford algebra.}
    \label{tab:parameters_linear}
\end{table}
\section{Experimental Results}\label{sec:exp}
We integrate our optimized backends into the existing Python package for Clifford neural layers implemented in the PyTorch libraries~\cite{paszkePyTorchImperativeStyle2019}. We use modern Python packaging and C++ PyTorch bindings to make the presence of the optimized backends transparent to the user. We achieved this by replacing the body of the forward function of the Clifford layers PyTorch modules with our optimized backends; this keeps the interface to the library unchanged. The whole package can be installed in a virtual environment with a single \verb|pip install .| command, as in the original implementation. We use the existing test suite to test the correctness of our backends and add additional tests to check the approximate numerical match of the outputs of our backends with the existing implementation.

\mypar{Experimental setup}
We ran our experiments using the AVX2 instruction set, but the instructions we used are compatible with all SIMD vector architectures.
We report the experiments run on an Intel Core i7-1280P, Alder Lake architecture, with a base frequency of performance cores of 1.8 GHz and efficient cores of 1.3 GHz. This architecture has two ports for floating point operations, and common multiplication and addition operations take four clock cycles~\cite{Intel64IA32}.
For cache sizes, the system has Golden Cove L1 Cache with 48 KB (Data), Golden Cove L2 Cache with 1.25 MB and a Total L3 Cache with 24 MB (3 MB × 8 Gracemont Cores).
We compile our code using a \texttt{gcc} version 13.3.0 compiler with flags \texttt{-O3}, \texttt{-march=native}, \texttt{-ffast-math}, \texttt{-fno-tree-vectorize}, \texttt{-mfma}, \texttt{-mavx2}.
We test the same functions over different input variables $n$ with \textit{warm} cache and all other parameters fixed. Fixed and variable parameters can be found in Tables \ref{tab:parameters_linear}, \ref{tab:parameters_act_linear} and  \ref{tab:parameters_conv}.
The total time to run all the experiments is around 30 minutes.

\begin{table}
    \centering
    \begin{tabular}{@{}lccccc@{}}
        \toprule
        Layer   & B & C & D & H & N\\
        \midrule
        Sum VSiLU    & $n$    & $n$   & -- & -- & --  \\
        Mean VSiLU   & $n$    & $n$   & -- & -- & --  \\
        Linear VSiLU & $n$    & $n$   & 20 & 20 & 3   \\
        \bottomrule
    \end{tabular}
    \caption{Parameters for activation layers, $n$ is input size parameter. 
    Activation functions apply vector SiLU to vector sum, mean and linear combination of vectors in G3.
    \textbf{Abbreviatons}: B=batch size, C=channels, D=depth, H=height, N=blades.}
    \label{tab:parameters_act_linear}
\end{table}

\begin{table}
    \centering
    \begin{tabular}{@{}lcccccccccc@{}}
        \toprule
        Layer        & CI & L$_1$ & N & CO & K & S & P & Di & G & B \\
        \midrule
        g3           & 8  & 10    & 3 & 10 & 10 & 1 & 0 & 1 & 1 & $n$ \\
        g3 trans     & 4  & 32    & 3 & 2  & 4  & 1 & 0 & 1 & 1 & $n$ \\
        1D          & 8  & 128   & 2 & 8  & 16 & 1 & 0 & 1 & 1 & $n$ \\
        2D        & 4  & 32    & 4 & 2  & 16 & 1 & 0 & 1 & 1 & $n$ \\
        3D        & 4  & 16    & 8 & 16 & 4  & 1 & 0 & 1 & 1 & $n$ \\
        \bottomrule
    \end{tabular}
    \caption{
    Parameters used in benchmarks for convolutional layers, $n$ is input size parameter
        \textbf{Abbreviations:} CI = input channels, L$_1$ = input size, N = number of blades, CO = output channels, K = kernel size, S = stride, P = padding, Di = dilation, G = groups, B = batch size.
    }
    \label{tab:parameters_conv}
\end{table}


\subsection{Results} \label{sec:results}

We now present the results obtained for the optimizations discussed in Section \ref{sec:optimizations} as well as the location of the algorithms in the roofline plot. Additionally, we provide a comparison with PyTorch.

\begin{figure}
    \centering 
    \includegraphics[width=0.5\textwidth]{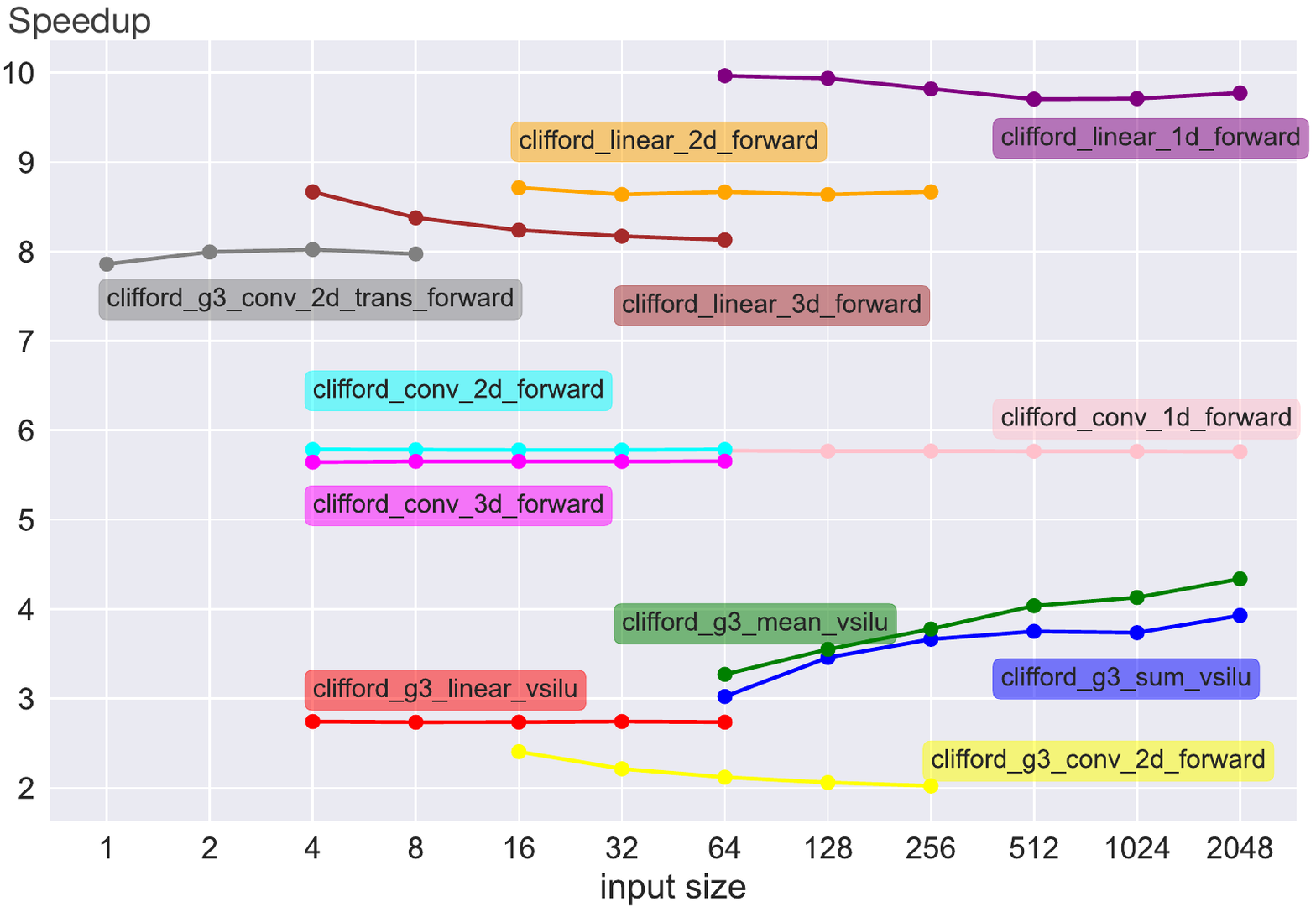}
    \caption{Speedup of each optimized function against their corresponding baseline function, without applying vectorization techniques. Corresponding input size parameter is stated in Tables \ref{tab:parameters_linear}, \ref{tab:parameters_act_linear} and  \ref{tab:parameters_conv} as $n$. }
    \label{fig:scalar_speedup}
\end{figure}

\mypar{Effects of optimizations} Analyzing Figure~\ref{fig:scalar_speedup}, we observe that the first optimization, based on its inherent mathematical structure and unrolling, has greatly improved most functions by an average of 6x. Noticeably, we remark a great improvement for the linear layers, while not equally great for the activation functions, and finally, intermediate results for the convolutional layers.

Convolution g3 forward has a different underlying kernel from the general Clifford kernel used in 2D convolution, so the results in optimizations vary between these two functions. Meanwhile, g3 convolution transposed greatly benefits from the inlining of the function and improving the access pattern.


\begin{figure}
    \centering 
    \includegraphics[width=0.50\textwidth]{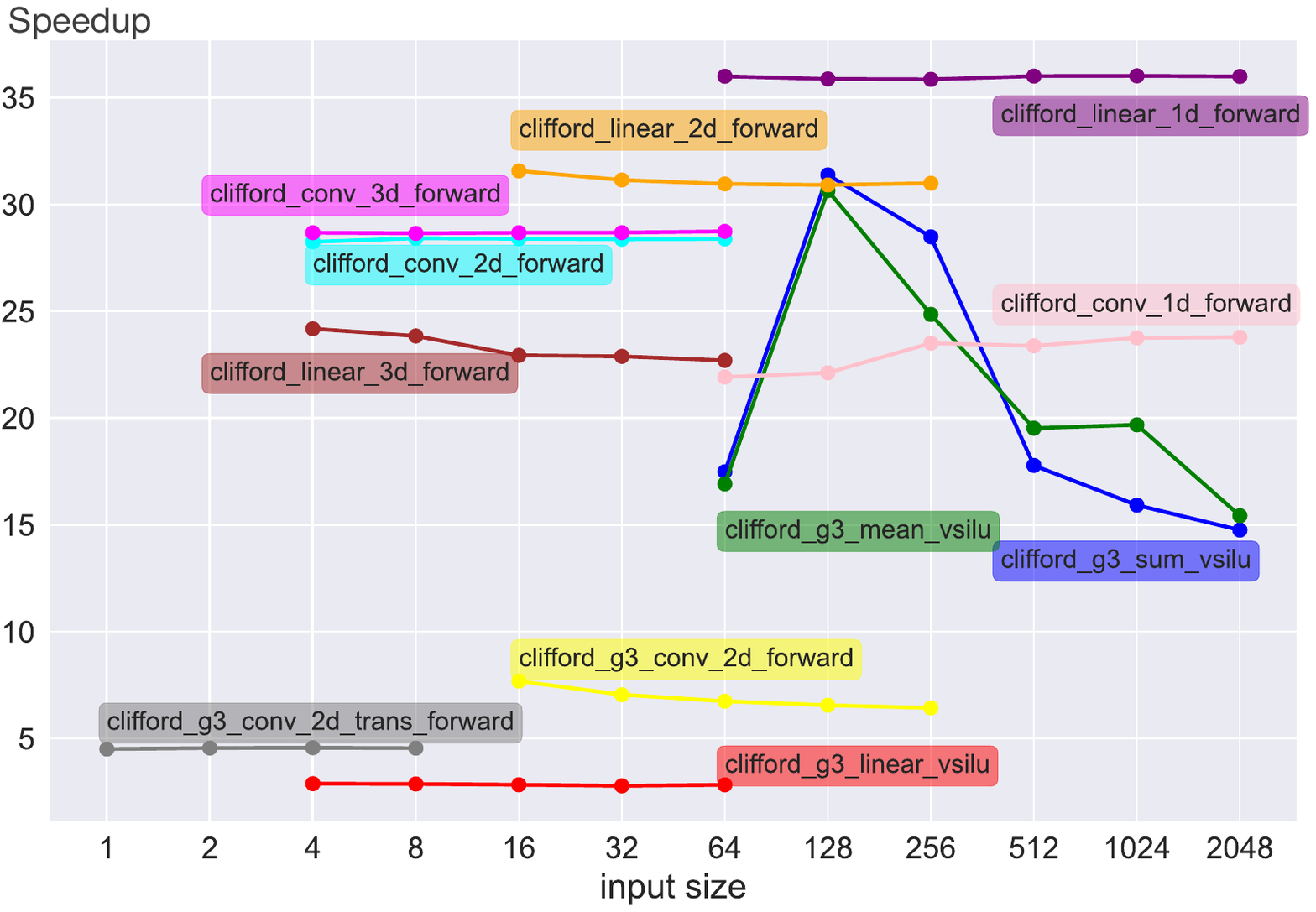}
    \caption{Speedup of each optimized function against their corresponding baseline function with SIMD instructions. We compute the average across all values reported in this plot and achieve overall average speed-up of 21.35 times over the baseline implementation.
    }
    \label{fig:vectorised_speedup}
\end{figure}

When looking at the results of the vectorization, shown in Figure~\ref{fig:vectorised_speedup}, we notice that the achieved speed-ups have increased by an average factor of 4 compared to the scalar optimizations. 
Lastly, the vectorization of Clifford g3 transposed convolution did not yield a runtime improvement. The combination of kernel computations and the convolution access pattern resulted in non-contiguous accesses in all directions when allowing for stride and padding greater than 1.

\mypar{The blessing of dimensionality} Contrary to the widely-held notion on the complexity of higher dimensions~\cite{koppenCurseDimensionality}, we observed that among the convolutional layers with standard kernel and dilation configurations, we were able to vectorize only the 3D convolution layer effectively since a multivector $\va$ from a 3-dimensional Clifford algebra has $2^3 = 8$ single-precision floating-point blades which fit perfectly inside an AVX2 register without the need of reordering, while we obtain worse results in 1D and 2D convolution since multivectors from $G_{1}$ and $G_{2}$ only have 2 and 4 single-precision floating-point blades, which do can fit more than one Clifford vector into a single 256-bit AVX2 register and therefore need additional operations to be in the correct format for computation, resulting in higher memory transfers and memory bottlenecks. We leverage this kind of optimization for 3D convolutional kernels, while for the other cases we assume a higher kernel size and absence of dilation and report those results.

\begin{figure}[!ht]
    \centering
    \begin{subfigure}[b]{0.47\textwidth}
        \centering
        \includegraphics[width=1\textwidth]{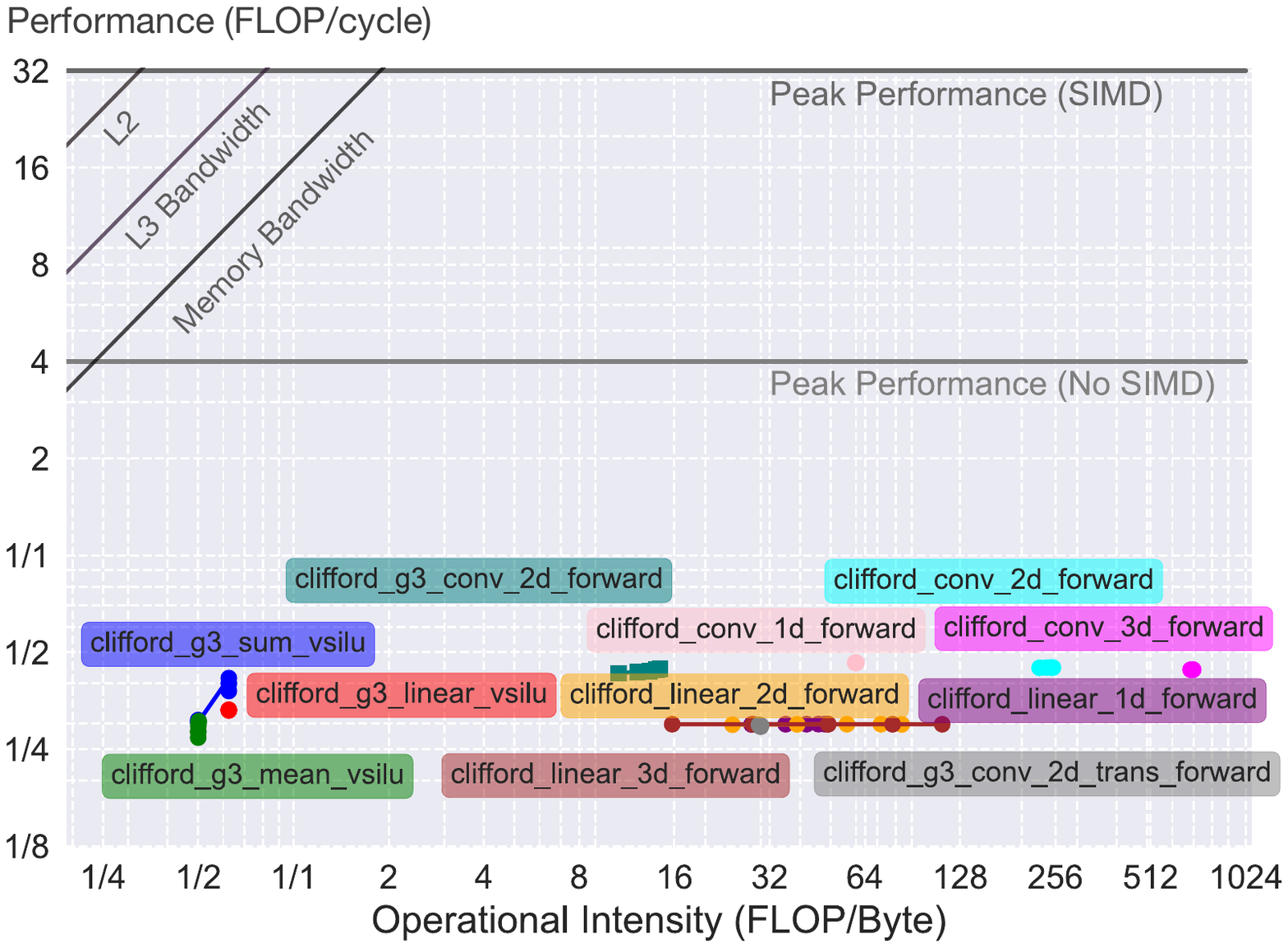}
        \caption{Roofline plot of the baseline version of each function.}
        \label{fig:roofline-base}
    \end{subfigure}
    \begin{subfigure}[b]{0.47\textwidth}
        \centering
        \includegraphics[width=1\textwidth]{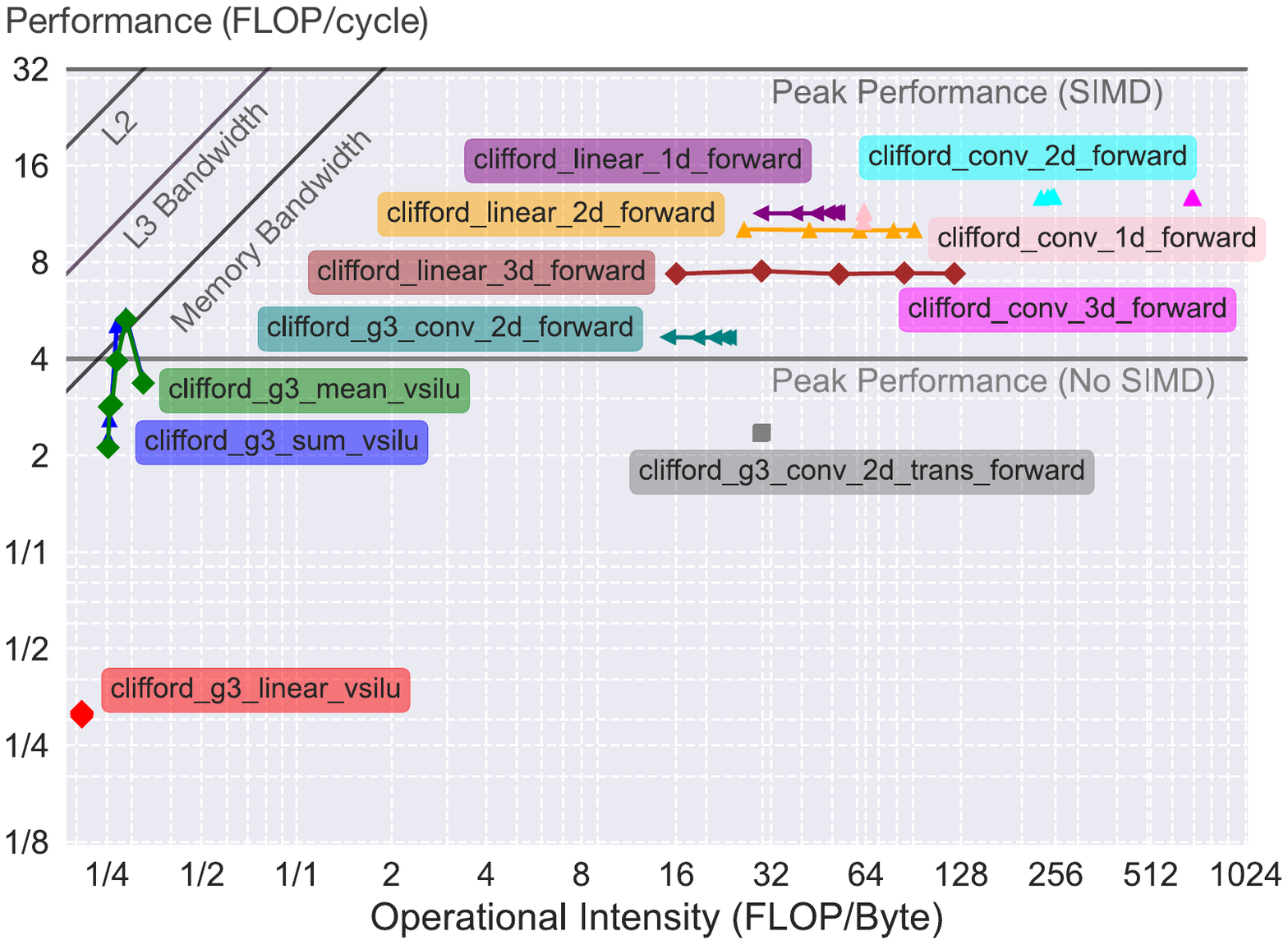}
        \caption{Roofline plot of the most performing version of each function. Note that \texttt{clifford\_g3\_linear\_vsilu} is notably bottlenecked by the exponential function, see~\ref{par:other-optimizations}.}
        \label{fig:roofline-opt}
    \end{subfigure}
    
    \caption{Roofline plots. Flops and bytes transferred were gathered using \texttt{perf} system call. Input size per function is the same as reported in Figure \ref{fig:scalar_speedup} and \ref{fig:vectorised_speedup}.}
    \label{fig:roofline}
\end{figure}

\mypar{Roofline analysis} Figure~\ref{fig:roofline} presents the roofline plot of the baseline and best-performing version of each function. We observe that all functions except for the activation functions are located in the compute-bound region. When comparing Figure~\ref{fig:roofline-opt} to ~\ref{fig:roofline-base}, we observe that the effectuated optimizations significantly improved the performance of most functions, often surpassing the peak performance achievable without SIMD.
However, this was not the case for \texttt{clifford\_g3\_linear\_vsilu}, where a bottleneck in the sigmoid function limited performance. We were unable to vectorize this function effectively without incurring an increase in runtime.


\mypar{Closing the gap with PyTorch} Finally, we test the performance of our custom kernels against the original repository written in PyTorch on a benchmark suite available in the project repository. A general overview of the speedups is presented in Table~\ref{tab:pytorch_comp}. We report three different classes of results: we (1) strongly outperform PyTorch with the transposed g3 convolutions (2) slightly outperform PyTorch in every activation function and some small inputs in the linear convolutional layers and (3) reach similar performance in terms of order of magnitude and half in terms of absolute performance for other functions. Overall, we delight ourselves to have beaten the original implementation.

\begin{table}[ht]
    \centering
    \begin{tabular}{@{}lc@{}}
        
        Function   & Speedup vs Pytorch  \\
        \toprule
clifford\_1d\_forward & 0.54x \\
clifford\_2d\_forward & 0.63x \\
clifford\_3d\_forward & 0.58x \\
clifford\_linear\_1d\_forward & \textbf{1.19x} \\
clifford\_linear\_2d\_forward & \textbf{1.00x} \\
clifford\_linear\_3d\_forward & 0.75x \\
clifford\_g3\_conv\_2d\_forward & 0.43x \\
clifford\_g3\_conv\_trans\_2d\_forward & \textbf{4.17x} \\
clifford\_g3\_linear\_vsilu & \textbf{1.26x} \\
clifford\_g3\_sum\_vsilu & \textbf{1.06x} \\
clifford\_g3\_mean\_vsilu & \textbf{1.05x} \\
        \bottomrule
    \end{tabular}
    \caption{Speedup of our best version for each function with respect to the original PyTorch implementation averaged over all tested inputs. We test for batch sizes equal to 16, 32, 64, 128 and 256.}
    \label{tab:pytorch_comp}
\end{table}

\section{Conclusions}
In this project, we effectively ported a Python implementation of the Clifford neural layers into C and optimized the most common layers, such as linear, convolutional and activation layers, for modern superscalar architectures equipped with AVX2 vector intrinsics. We discovered mathematical properties of the algebra that, when applied, greatly enhance the performance over the baseline implementation.
Overall, we achieve on-par performance with PyTorch in most functions, occasionally surpassing and lagging shortly behind in a few other cases.
\bibliographystyle{IEEEbib}
\bibliography{bibl_conf}






\end{document}

%% file: algorithms/cliffordlinearbasline.tex
\begin{algorithm}
    \caption{Clifford linear layer in Cl$_{1,0}(\mathbb{R})$.}
    \label{app:algo_clifford_linear}
    \begin{mdframed}[backgroundcolor=gray!5,rightline=false,leftline=false]
        \begin{algorithmic}[1]
            \Function{CliffordLinearKernel1d}{$\mW$}
            \State $(O, I, 2) \gets$ \Call{Shape}{$\mW$}
            \State kernel $\gets \begin{bmatrix}
                \mW[:, :,0] & -\mW[:, :, 1] \\
                \mW[:, :, 1] & \mW[:, :, 0] \\
            \end{bmatrix}$
            \State \Return{kernel}
            \EndFunction
            \Function{CliffordLinear1D}{$\mW$, $\vb$, $\mX$}
                \State $(O, I, 2) \gets$ \Call{Shape}{$\mW$}
                \State $(O, 2) \gets$ \Call{Shape}{$\vb$}
                \State $(B, I, 2) \gets$ \Call{Shape}{$\mX$}
                \State weight $\gets$ \Call{CliffordLinearKernel1D}{$\mW$}
                \State bias $\gets$ \Call{ViewAsRealTensor}{$\vb$}
                \State bias $\gets$ \Call{Permute}{$\vb$, (1, 0)}
                \State bias $\gets$ \Call{Flatten}{bias}
                \State input $\gets$ \Call{ViewAsRealTensor}{$\mX$}
                \State input $\gets$ \Call{Permute}{input, (2, 0, 1)}
                \State input $\gets$ \Call{Flatten}{input}
                \State output $\gets$ \Call{Linear1D}{weight, bias, input}
                \State output $\gets$ \Call{Reshape}{(2, O)}
                \State output $\gets$ \Call{Permute}{output, (0, 1)}
                \State \Return{\Call{ViewAsMultivector}{output}}
            \EndFunction
        \end{algorithmic}
    \end{mdframed}
\end{algorithm}

%% file: algorithms/cliffordlinearoptimized.tex
\begin{algorithm}
    \caption{Optimized Clifford linear layer in Cl$_{1,0}(\mathbb{R})$.}
    \label{app:algo_optimized_clifford_linear}
    \begin{mdframed}[backgroundcolor=gray!5,rightline=false,leftline=false]
        \begin{algorithmic}[1]
            \Function{OptimizedCliffordLinear1D}{$\mW$, $\vb$, $\mX$}
                \State $(O, I, 2) \gets$ \Call{Shape}{$\mW$}
                \State $(B, I, 2) \gets$ \Call{Shape}{$\mX$}
                \State $(O, 2) \gets$ \Call{Shape}{$\vb$}
                \State $\mY \gets$ \Call{Zeros}{$(B, O, 2)$}
                \For{$b = 0$ to $B - 1$}
                    \For{$o = 0$ to $O - 1$}
                        \State $\text{sum}_{00}, \text{sum}_{01}, \text{sum}_{10}, \text{sum}_{11} \gets 0$
                        \For{$i = 0$ to $I - 1$}
                            \State $\text{sum}_{00} \mathrel{+}= \mW[o, i, 0] \cdot \mX[b, i, 0]$
                            \State $\text{sum}_{01} \mathrel{+}= \mW[o, i, 0] \cdot \mX[b, i, 1]$
                            \State $\text{sum}_{10} \mathrel{+}= \mW[o, i, 1] \cdot \mX[b, i, 0]$
                            \State $\text{sum}_{11} \mathrel{+}= \mW[o, i, 1] \cdot \mX[b, i, 1]$
                        \EndFor
                        \State $\mY[b, o, 0] \gets \text{sum}_{00} - \text{sum}_{11} + \vb[o, 0]$
                        \State $\mY[b, o, 1] \gets \text{sum}_{01} + \text{sum}_{10} + \vb[o, 1]$
                    \EndFor
                \EndFor
                \State \Return{$\mY$}
            \EndFunction
        \end{algorithmic}
    \end{mdframed}
\end{algorithm}